\documentclass[
 twocolumn,
 superscriptaddress,
 groupedaddress,
 showpacs,
 amsmath,
 amssymb,
 aps,
 prl,
]{revtex4-1}

\usepackage[dvipdfmx]{graphicx}% Include figure files
\usepackage{dcolumn}% Align table columns on decimal point
\usepackage{bm}% bold math

\usepackage{algcompatible}
\usepackage[size=small]{caption}
\DeclareCaptionFormat{algorithm}{\hrulefill\vskip-8pt\hrulefill\par#1#2#3\vskip-4pt\hrulefill}
\usepackage{etoolbox}
\AtEndEnvironment{algorithm}{\noindent\hrulefill\par\nobreak\vskip-8pt\hrulefill}
\usepackage{newfloat}
\DeclareFloatingEnvironment[
  fileext=loa,
  listname=List of Algorithms,
  name=ALGORITHM,
  placement=tbhp,
]{algorithm}
\DeclareMathOperator*{\argmax}{arg\,max}
\usepackage{mathtools}
\usepackage{amsthm}

\usepackage{subcaption}
\usepackage{xcolor}
\usepackage{diagbox}
\usepackage{multirow}
\usepackage{xr}
\usepackage{braket}
\usepackage{amsmath}

\makeatletter
\newcommand{\vast}{\bBigg@{3.0}}
\newcommand{\Vast}{\bBigg@{4.0}}

\makeatother
\begin{document}

\preprint{APS/123-QED}

% \input{paper-dqavb-spin-01-01-title.tex}
% !TEX root = ./paper-dqavb-spin-00-01-main.tex
\title{A Quantum Extension of Variational Bayes Inference}% Force line breaks with \\
% \title{Deterministic quantum annealing variational Bayes inference}% Force line breaks with \\
%\thanks{confidential}%

\author{Hideyuki Miyahara$^1$}
\email{hideyuki\_miyahara@mist.i.u-tokyo.ac.jp}
\author{Yuki Sughiyama$^2$}
% \email{hmiyahara512@gmail.com}

% \altaffiliation[Also at ]{Department of Information and Technology, the University of Tokyo.}%Lines break automatically or can be forced with \\
\affiliation{%
$^1$Department of Mathematical Informatics,
Graduate School of Information Science and Technology,
The University of Tokyo,
7-3-1 Hongosanchome Bunkyo-ku Tokyo 113-8656, Japan
% 4-6-1, Komaba, Meguro-ku, Tokyo 153-8505, Japan
}%

\affiliation{%
$^2$Institute of Industrial Science, The University of Tokyo,
4-6-1, Komaba, Meguro-ku, Tokyo 153-8505, Japan
}%

\date{\today}% It is always \today, today,

% \input{paper-dqavb-spin-02-01-abst.tex}
% !TEX root = ./paper-dqavb-spin-00-01-main.tex

\begin{abstract}
Variational Bayes (VB) inference is one of the most important algorithms in machine learning and widely used in engineering and industry.
However, VB is known to suffer from the problem of local optima.
In this Letter, we generalize VB by using quantum mechanics, and propose a new algorithm, which we call quantum annealing variational Bayes (QAVB) inference.
We then show that QAVB drastically improve the performance of VB by applying them to a clustering problem described by a Gaussian mixture model.
% We then show that, by applying it to a clustering problem described by a Gaussian mixture model, QAVB with specific annealing schedules drastically improve the performance of VB.
Finally, we discuss an intuitive understanding on how QAVB works well.
\end{abstract}

%

% \pacs{Valid PACS appear here}
% \pacs{02.50.Fz, 02.50.Tt, 02.60.Pn} % First one
% \pacs{05.30.-d, 75.10.Nr, 89.70.+c} % Kadowaki and Nishimori
% \pacs{75.10.Nr, 03.67.Ac, 64.70.Tg, 75.50.Lk} % https://arxiv.org/pdf/0806.4144.pdf
% \pacs{03.67.-a, 03.67.Ac, 89.90.+n, 89.70.-a, 89.20.-a}

\maketitle

% \tableofcontents

% \input{paper-dqavb-spin-03-01-intro.tex}
% !TEX root = ./paper-dqavb-spin-00-01-main.tex
% \section{Introduction}

\textit{Introdction.}---
Machine learning gathers considerable attention in a wide range of fields, and much effort is devoted to develop effective algorithms.
Variational Bayes (VB) inference~\cite{Waterhouse01, Attias01, Jordan01, Blei01, Bishop01, Murphy01} is one of the most fundamental methods in machine learning, and widely used for parameter estimation and model selection.
In particular, VB has succeeded to compensate some disadvantages of the expectation-maximization (EM) algorithm~\cite{Dempster01, Bishop01, Murphy01}, which is a well-used approach for maximum likelihood estimation.
For example, overfitting, which is often occurred in EM, is greatly moderated in VB.
Furthermore, a variant of VB based on classical statistical mechanics, which we call simulated annealing variational Bayes (SAVB) inference in this paper, was proposed~\cite{Katahira01} and has been getting popular in many fields due to its effectiveness.
However, it is also known that VB and SAVB often fail to estimate appropriate parameters of an assumed model depending on prior distributions and initial conditions.

In the field of physics, the study of quantum computation and how to exploit it for machine learning are getting popular.
For example, while experimentalists are intensively developing quantum machines~\cite{Barends01, Mohseni01, Johnson01, Johnson02, Albash02}, theorists are developing quantum error correction schemes~\cite{Devitt01, Bennett01, Gottesman01, Knill01, Pudenz01} and quantum algorithms~\cite{Harrow01, Robentrost01, Lloyd01, Apolloni01, Finnila01, Kadowaki01, Farhi01, Albash01, Miyahara03, Miyahara04, Miyahara05}.
In particular, the study of quantum annealing (QA) has a history for more than two decades~\cite{Apolloni01, Finnila01, Kadowaki01, Farhi01} and is still progressing~\cite{Albash01}.

In this Letter, by focusing on QA and VB, we devise a quantum-mechanically inspired algorithm that works on a classical computer in practical time and achieves a considerable improvement over VB and SAVB.
More specifically, we introduce the mathematical mechanism of quantum fluctuations into VB, and propose a new algorithm, which we call quantum annealing variational Bayes (QAVB) inference.
To demonstrate the performance of QAVB, we consider a clustering problem and employ a Gaussian mixture model, which is one of important applications of VB.
Then, we see that QAVB succeeds in estimation with high probability while VB and SAVB do not.
This fact is noteworthy because our algorithm is one of the few algorithms that can obtain a global optimum of non-convex optimization in practical computational time without using random numbers.
%

% \input{paper-dqavb-spin-04-02-vb.tex}
% !TEX root = ./paper-dqavb-spin-00-01-main.tex
% \section{Variational Bayes inference} \label{vb-55}

\textit{Problem setting of VB.}---
For preparation of a quantum extension of VB, we briefly review the problem setting of VB~\cite{Waterhouse01, Attias01, Jordan01, Blei01, Bishop01, Murphy01}.
First, we summarize the definitions of variables.
Suppose that we have $N$ data points $Y^\mathrm{obs} = \{y_{i}^\mathrm{obs}\}_{i=1}^N$, which are independent and identically distributed by the conditional distribution $p^{y, \sigma| \theta} (y_i, \sigma_i| \theta)$, where $y_i$, $\sigma_i$, and $\theta$ are an observable variable, a hidden variable and a parameter, respectively.
Thus, we have $p^{Y, \Sigma| \theta} (Y, \Sigma| \theta) = \prod_{i=1}^N p^{y, \sigma| \theta} (y_i, \sigma_i| \theta)$, where $Y = \{ y_i \}_{i=1}^N$ and $\Sigma = \{ \sigma_i \}_{i=1}^N$.
The joint distribution is also given by $p^{Y, \Sigma, \theta} (Y, \Sigma, \theta) = p^{Y, \Sigma, \theta} (Y, \Sigma| \theta) p_\mathrm{pr}^\theta (\theta)$, where $p_\mathrm{pr}^\theta (\theta)$ denotes the prior distribution of $\theta$.
Furthermore, we define the domains of $\Sigma$ and $\theta$ as $S^\Sigma \coloneqq \overset{N}{\underset{i=1}{\otimes}} S^\sigma$ and $S^\theta$, respectively.

The goal of VB is to approximate the posterior distributions given by $p^{\Sigma, \theta| Y}(\Sigma, \theta| Y^\mathrm{obs}) = p^{Y, \Sigma, \theta} (Y^\mathrm{obs}, \Sigma, \theta) / p^Y (Y^\mathrm{obs})$ with $p^Y (Y^\mathrm{obs}) = \sum_{\Sigma \in S^\Sigma} \int_{\theta \in S^\theta} d\theta \, p(Y^\mathrm{obs}, \Sigma, \theta)$ in the mean field approximation.
Here, we have used the Bayes theorem for the derivation of the posterior distribution.
Using a variational function $q^{\Sigma, \theta}(\Sigma, \theta)$ that satisfies $\sum_{\Sigma \in S^\Sigma} \int_{\theta \in S^\theta} d\theta \, q^{\Sigma, \theta}(\Sigma, \theta) = 1$, the objective function of VB is given by
\begin{align}
& \mathrm{KL} \left( q^{\Sigma, \theta} (\cdot, \cdot) \middle\| p^{\Sigma, \theta | Y} (\cdot, \cdot | Y^\mathrm{obs}) \right) \nonumber \\
& \quad \coloneqq - \sum_{\Sigma \in S^\Sigma} \int_{\theta \in S^\theta} d\theta \, q^{\Sigma, \theta} (\Sigma, \theta) \ln \frac{p^{\Sigma, \theta | Y} (\Sigma, \theta | Y^\mathrm{obs})}{q^{\Sigma, \theta} (\Sigma, \theta)}, \label{KL-VB-01}
\end{align}
which is the KL divergence~\cite{Kullback01, Kullback02}.
In VB, we minimize Eq.~\eqref{KL-VB-01} in the mean field approximation given by
\begin{align}
q^{\Sigma, \theta} (\Sigma, \theta) &= q^\Sigma (\Sigma) q^\theta (\theta). \label{mean-field-approximation-01}
\end{align}
Thus, by setting the functional derivatives of Eq.~\eqref{KL-VB-01} under Eq.~\eqref{mean-field-approximation-01} with respect to $q^\Sigma (\Sigma)$ and $q^\theta(\theta)$ equal to 0 and solving for $q^\Sigma (\Sigma)$ and $q^{\theta} (\theta)$, we obtain the update equations for $\Sigma$ and $\theta$:
\begin{align}
q_{t+1}^{\Sigma} (\Sigma) &\propto \exp \bigg( \int_{\theta \in S^\theta} d\theta\, q_{t+1}^\theta(\theta) \ln p^{Y, \Sigma, \theta} (Y^\mathrm{obs}, \Sigma, \theta) \bigg), \label{VB-Estep01} \\
q_{t+1}^{\theta} (\theta) &\propto \exp \Bigg( \sum_{\Sigma \in S^{\Sigma}} q_t^{\Sigma}(\Sigma) \ln p^{Y, \Sigma, \theta} (Y^\mathrm{obs}, \Sigma, \theta) \Bigg), \label{VB-Mstep01}
\end{align}
where  $q_t^\Sigma (\Sigma)$ and $q_t^\theta(\theta)$ is the distributions of $\Sigma$ and $\theta$ at the $t$-th iteration~\cite{Bishop01, Murphy01}.

VB is widely used due to its effectiveness.
In some cases, the performance of VB is much better than that of EM~\cite{Bishop01, Murphy01, Dempster01}, and VB can be directly used for model selection~\cite{Waterhouse01, Attias01, Jordan01, Blei01, Bishop01, Murphy01}.
However, it is also known that the performance of VB heavily depends on initial conditions.
To relax this problem, we introduce quantum fluctuations to VB in the rest of this Letter.
%

% \input{paper-dqavb-spin-05-01-dqavb.tex}
% !TEX root = ./paper-dqavb-spin-00-01-main.tex
% \section{Quantum annealing variational Bayes inference} \label{DQAVB-00}

\textit{Quantum annealing variational Bayes inference.}---
Here, we formulate a quantum extension of VB.
We first define the classical Hamiltonians by $p^{Y, \Sigma| \theta} (Y^\mathrm{obs}, \Sigma| \theta)$ and $p_\mathrm{pr}^\theta (\theta)$:
\begin{align}
H_\mathrm{cl}^{\Sigma| \theta} &\coloneqq - \ln p^{Y, \Sigma| \theta} (Y^\mathrm{obs}, \Sigma| \theta), \label{classical-hamiltonian-11} \\
H_\mathrm{pr}^\theta &\coloneqq - \ln p_\mathrm{pr}^{\theta} (\theta). \label{classical-hamiltonian-12}
\end{align}
Next, we define operators $\hat{\sigma}_i$ and $\hat{\theta}$ whose eigenvalues are $\sigma_i$ and $\theta$, respectively; that is, $\hat{\sigma}_i$ and $\hat{\theta}$ satisfy
$\hat{\sigma}_i \Ket{\sigma_i} = \sigma_i \Ket{\sigma_i}$ and $\hat{\theta} \Ket{\theta} = \theta \Ket{\theta}$, where $\Ket{\sigma_i}$ and $\Ket{\theta}$ are eigenstates of $\hat{\sigma}_i$ and $\hat{\theta}$, respectively.
In this paper, we assume $\hat{\sigma}_i$ and $\hat{\theta}$ are commutative with each other.
Using the above definition of $\Ket{\sigma_i}$, we also define $\Ket{\Sigma} \coloneqq \overset{N}{\underset{i=1}{\otimes}} \Ket{\sigma_i}$.
Then, we replace $\Sigma = \{ \sigma_i \}_{i=1}^N$ and $\theta$ in Eqs.~\eqref{classical-hamiltonian-11} and \eqref{classical-hamiltonian-12} by $\bigg\{ \bigg( \overset{i-1}{\underset{j = 1}{\otimes}} \hat{I}^{\sigma_j} \bigg) \otimes \hat{\sigma}_i \otimes \bigg( \overset{N}{\underset{j = i+1}{\otimes}} \hat{I}^{\sigma_j} \bigg) \bigg\}_{i=1}^N$ and $\hat{\theta}$, respectively, where $\hat{I}^{\sigma_i}$ denotes the identity operator for the spaces spanned by $\Ket{\sigma_i}$.
That is, we define
\begin{align}
\hat{H}_\mathrm{cl}^{\Sigma| \theta} &\coloneqq \sum_{\Sigma \in S^\Sigma} \int_{\theta \in S^\theta} d \theta \, H_\mathrm{cl}^{\Sigma| \theta} \hat{P}^{\Sigma, \theta}, \label{classical-hamiltonian-10-02} \\
\hat{H}_\mathrm{pr}^{\theta} &\coloneqq \sum_{\Sigma \in S^\Sigma} \int_{\theta \in S^\theta} d \theta \, H_\mathrm{pr}^{\theta} \hat{P}^{\Sigma, \theta},
\end{align}
where $\hat{P}^{\Sigma, \theta} \coloneqq \hat{P}^\Sigma \otimes \hat{P}^\theta$, $\hat{P}^\Sigma \coloneqq \overset{N}{\underset{i=1}{\otimes}} \hat{P}^{\sigma_i}$, $\hat{P}^{\sigma_i} \coloneqq \Ket{\sigma_i} \Bra{\sigma_i}$, and $\hat{P}^\theta \coloneqq \Ket{\theta} \Bra{\theta}$.
To introduce quantum fluctuations to VB, we define a Gibbs operator that involves a non-commutative term:
\begin{align}
\hat{f} (\beta, s) &\coloneqq \exp \left( - \hat{H}_\mathrm{pr}^\theta - \beta (1 - s) \hat{H}_\mathrm{cl}^{\Sigma| \theta} - \beta s \hat{H}_\mathrm{qu}^\Sigma \right), \label{Gibbs-05-01}
\end{align}
where $\hat{H}_\mathrm{qu}^\Sigma$ is a non-commutative term, defined as $\hat{H}_\mathrm{qu}^\Sigma \coloneqq \sum_{i=1}^N \hat{H}_\mathrm{qu}^{\sigma_i}$, and $\hat{H}_\mathrm{qu}^{\sigma_i}$ is defined such that
\begin{align}
\bigg[ \hat{H}_\mathrm{qu}^{\sigma_i}, \bigg( \overset{i-1}{\underset{j = 1}{\otimes}} \hat{I}^{\sigma_j} \bigg) \otimes \hat{\sigma}_i \otimes \bigg( \overset{N}{\underset{j = i+1}{\otimes}} \hat{I}^{\sigma_j} \bigg) \otimes \hat{I}^\theta \bigg] \ne 0, \label{quantization-condition-01}
\end{align}
for any $i$~\footnote{In Eq.~\eqref{Gibbs-05-01}, we intentionally drop $\beta$ from the term including $\hat{H}_\mathrm{pr}^\theta$. The reason is that, when $\beta$ is large, a necessary condition of a conjugate prior distribution may be broken. In the case of a GMM, large $\beta$ breaks a condition of the Wishart distribution, which is the conjugate prior distribution for the inverse of the covariance of a Gaussian function.}.
Here, $\hat{I}^\theta$ represents the identity operator for the space spanned by $\Ket{\theta}$.
This Gibbs operator, Eq.~\eqref{Gibbs-05-01}, involves two annealing parameters $\beta$ and $s$, where, in terms of physics, $\beta$ is regarded as the inverse temperature and $s$ represents the strength of quantum fluctuations.
Thus, when $s = 0$ and $\beta = 1$, we recover $\Braket{\Sigma, \theta | \hat{f} (\beta = 1, s = 0) | \Sigma, \theta} = p^{Y, \Sigma, \theta} (Y^\mathrm{obs}, \Sigma, \theta)$.
Although we consider only the quantization of $\Sigma$, the quantization of $\theta$ is almost straightforward~\footnote{An approach to quantize $\theta$ is just to add $\hat{H}_\mathrm{qu}^\theta$ that satisfies $[\hat{H}_\mathrm{qu}^\theta, \hat{I}^\Sigma \otimes \hat{\theta}] \ne 0$ to Eq.~\eqref{Gibbs-05-01}.}.

Using Eq.~\eqref{Gibbs-05-01}, we define a quantum extension of the KL divergence~\cite{Umegaki01} by
\begin{align}
&\mathcal{S} \left( \hat{\rho}^{\Sigma, \theta} \middle \| \frac{\hat{f} (\beta, s)}{\mathcal{Z} (\beta, s)} \right) \nonumber \\
& \quad \coloneqq - \mathrm{Tr}_{\Sigma, \theta} \bigg[ \hat{\rho}^{\Sigma, \theta} \bigg\{ \ln \frac{\hat{f} (\beta, s)}{\mathcal{Z} (\beta, s)} - \ln \hat{\rho}^{\Sigma, \theta} \bigg\} \bigg], \label{KL-DQAVB-01}
\end{align}
where $\mathcal{Z} (\beta, s) \coloneqq \mathrm{Tr}_{\Sigma, \theta} \left[ \hat{f} (\beta, s) \right]$ and $\mathrm{Tr}_{\Sigma, \theta} [\cdot] \coloneqq \sum_{\Sigma \in S^\Sigma} \int_{\theta \in S^\theta} d\theta \, \Braket{ \Sigma, \theta | \cdot | \Sigma, \theta }$.
Also, $\hat{\rho}^{\Sigma, \theta}$ denotes a density operator over $\Sigma$ and $\theta$ that satisfies $\mathrm{Tr}_{\Sigma, \theta} \left[ \hat{\rho}^{\Sigma, \theta} \right] = 1$.
In particular, when $\beta = 1$, $s = 0$, and $\hat{\rho}$ is diagonal, the quantum relative entropy, Eq.~\eqref{KL-DQAVB-01}, reduces to the classical KL divergence, Eq.~\eqref{KL-VB-01}.

To derive the update equations, we repeat the almost same procedure of VB; that is, we employ the mean field approximation $\hat{\rho}^{\Sigma, \theta} = \hat{\rho}^\Sigma \otimes \hat{\rho}^{\theta}$, where $\hat{\rho}^\Sigma$ and $\hat{\rho}^\theta$ represent the density operators for $\Sigma$ and $\theta$,respectively; then Eq.~\eqref{KL-DQAVB-01} can be reduced to~\footnote{See Sec.~A in the Supplemental Material for the detail derivation.}
\begin{align}
\mathcal{S} &\left( \hat{\rho}^\Sigma \otimes \hat{\rho}^{\theta} \middle \| \frac{\hat{f} (\beta, s)}{\mathcal{Z} (\beta, s)} \right) \nonumber \\
& = - \sum_{\Sigma \in S^\Sigma} \sum_{\Sigma' \in S^\Sigma} \int_{\theta \in S^\theta} d\theta \, \int_{\theta' \in S^\theta} d\theta' \, \nonumber \\
& \qquad \times \Braket{\Sigma| \hat{\rho}^\Sigma | \Sigma'} \Braket{\theta | \hat{\rho}^\theta | \theta'} \nonumber \\
& \qquad \times \Big[ \Bra{\Sigma'} \otimes \Bra{\theta'} \Big] \Big[ \ln \hat{f} (\beta, s) \Big] \Big[ \Ket{\Sigma} \otimes \Ket{\theta} \Big] \nonumber \\
& \quad + \sum_{\Sigma \in S^\Sigma} \sum_{\Sigma' \in S^\Sigma} \Bra{\Sigma} \hat{\rho}^{\Sigma} \Ket{\Sigma'} \Bra{\Sigma'} \ln \hat{\rho}^{\Sigma} \Ket{\Sigma} \nonumber \\
& \quad + \int_{\theta \in S^\theta} d\theta \, \int_{\theta' \in S^\theta} d\theta' \, \Bra{\theta} \hat{\rho}^{\theta} \Ket{\theta'} \Bra{\theta'} \ln \hat{\rho}^{\theta} \Ket{\theta} \nonumber \\
& \quad + \ln \mathcal{Z} (\beta, s). \label{free-mean-42}
\end{align}
Next, by setting the functional derivatives of Eq.~\eqref{free-mean-42} with respect to $\Braket{\Sigma | \hat{\rho}^\Sigma | \Sigma'}$ and $\Braket{\theta | \hat{\rho}^{\theta} | \theta'}$ equal to 0 and solving for $\hat{\rho}^\Sigma$ and $\hat{\rho}^\theta$, we obtain the update equations~\footnote{See Sec.~B in the Supplemental Material for the detail derivation.}:
\begin{align}
\hat{\rho}_{t+1}^{\Sigma} &\propto \exp \left( \mathrm{Tr}_{\theta} \left[ \hat{\rho}_{t+1}^\theta \ln \hat{f} (\beta, s) \right] \right), \label{DQAVB-Estep01} \\
\hat{\rho}_{t+1}^{\theta} &\propto \exp \left( \mathrm{Tr}_{\Sigma} \left[ \hat{\rho}_t^\Sigma \ln \hat{f} (\beta, s) \right] \right), \label{DQAVB-Mstep01}
\end{align}
where $\mathrm{Tr}_{\Sigma} [\cdot] \coloneqq \sum_{\Sigma \in S^\Sigma} \Braket{ \Sigma | \cdot | \Sigma }$, $\mathrm{Tr}_{\theta} [\cdot] \coloneqq \int_{\theta \in S^\theta} d\theta \, \Braket{ \theta | \cdot | \theta }$,
and $t$ stands for the number of iterations.
We mention that $\mathrm{Tr}_{\Sigma} [\cdot]$ and $\mathrm{Tr}_{\theta} [\cdot]$ represent partial traces, and they yield operators on the spaces spanned by $\Ket{\theta}$ and $\Ket{\Sigma}$, respectively.
We also note that the subscriptions $t$ and $t+1$ in the right-hand sides of Eqs.~\eqref{DQAVB-Estep01} and \eqref{DQAVB-Mstep01} depend on implementations of QAVB and the normalization factors of Eqs.~\eqref{DQAVB-Estep01} and \eqref{DQAVB-Mstep01} are determined by the condition of density operators $\mathrm{Tr}_\Sigma [\hat{\rho}^\Sigma] = 1$ and $\mathrm{Tr}_\theta [\hat{\rho}^\theta] = 1$.
In QAVB, we iterate these two update equations changing the annealing parameters $\beta$ and $s$ until a termination condition is satisfied.
In this algorithm, we obtain density operators $\hat{\rho}_t^\Sigma$ and $\hat{\rho}_t^\theta$ in each step, and their diagonal elements $\Braket{\Sigma | \hat{\rho}_t^\Sigma | \Sigma}$ and $\Braket{\theta | \hat{\rho}_t^\theta | \theta}$ represent the distributions of $\Sigma$ and $\theta$, respectively.
In practical applications, we may use the mean $\mathrm{Tr}_\theta [\hat{\theta} \hat{\rho}^\theta]$ or the mode $\argmax_\theta \Braket{\theta | \hat{\rho}^\theta | \theta}$.
Note that, when $\beta = 1$ and $s = 0$, Eqs.~\eqref{DQAVB-Estep01} and \eqref{DQAVB-Mstep01} exactly reduces to the update equations of VB, Eqs.~\eqref{VB-Estep01} and \eqref{VB-Mstep01}.
Finally, we summarize this algorithm in Algo.~\ref{algo-DQAVB-04}.
\begin{algorithm}[t]
\caption{Quantum annealing variational Bayes (QAVB) inference} \label{algo-DQAVB-04}
\begin{algorithmic}[1]
\STATE set $\hat{\rho}_\mathrm{pr}^\theta$ and $t \leftarrow 0$, and initialize $\hat{\rho}_0^\Sigma$
\STATE set $\beta \leftarrow \beta_0$ and $s \leftarrow s_0$
\WHILE{convergence criterion is not satisfied}
\STATE compute $\hat{\rho}_{t+1}^\theta$ in Eq.~\eqref{DQAVB-Mstep01}
\STATE compute $\hat{\rho}_{t+1}^\Sigma$ in Eq.~\eqref{DQAVB-Estep01} for $i=1, 2, \dots, N$
\STATE change $\beta$ and $s$
\STATE $t \leftarrow t+1$
\ENDWHILE
\end{algorithmic}
\end{algorithm}

% \input{paper-dqavb-spin-05-02-dqavb.tex}
% \input{paper-dqavb-spin-05-03-dqavb.tex}
% \input{paper-dqavb-spin-05-04-dqavb.tex}
% \input{paper-dqavb-spin-05-05-dqavb.tex}

% \input{paper-dqavb-spin-06-01-model.tex}
% !TEX root = ./paper-dqavb-spin-00-01-main.tex
% \section{Gaussian mixture models} \label{gmm-01}

\textit{Gaussian mixture models.}---
To see the performance of QAVB, we consider the estimation problem of the parameters and number of clusters of a GMM studied in Ref.~\cite{Attias01, Bishop01, Murphy01}.
The joint probability distribution of the GMM over an observable variable $y_i$ and a hidden variable $\sigma_i$ conditioned by a set of parameters $\theta$ is given by
\begin{align}
p^{y, \sigma| \theta} (y_i, \sigma_i| \theta) &= \sum_{k=1}^K \pi^k \mathcal{N} (y_i| \mu^k, (\Lambda^k)^{-1}) \delta_{k, \sigma_i}, \label{joint-MM-01}
\end{align}
where $\delta_{k, \sigma_i}$ is the Kronecker delta function, $\{ \pi_k \}_{k=1}^K$ are the mixing coefficients for the GMM, and $\mathcal{N} (y_i| \mu^k, (\Lambda^k)^{-1})$ is a Gaussian distribution whose mean and precision, which is the inverse of covariance, are $\mu^k$ and $\Lambda^k$, respectively~\footnote{We have not got into the detail of the prior distribution of the GMM $p_\mathrm{pr}^\theta (\theta)$, because we do not quantize it in this paper. See Ref.~\cite{Bishop01, Murphy01}, if the reader is not familiar with it.}.
Here, we have assumed that each hidden variable $\sigma_i$ takes $1, \dots, K$; that is, $S^\sigma = \{k\}_{k=1}^K$~\footnote{When we use the one-hot notation~\cite{Bishop01, Murphy01}, we can construct an equivalent quantization scheme.}.
To simplify the notation, we denote $\{ \pi^k \}_{k=1}^K$, $\{ \mu^k \}_{k=1}^K$, and $\{ \Lambda^k \}_{k=1}^K$ by $\pi$, $\mu$, and $\Lambda$, respectively, and we refer by $\theta$ to $\{\pi, \mu, \Lambda\}$ collectively.

Taking the logarithm of Eq.~\eqref{joint-MM-01}, we define the Hamiltonian of the GMM for $\sigma_i$ with $y = y_i^\mathrm{obs}$ as
\begin{align}
H_\mathrm{cl}^{\sigma_i| \theta} &= - \ln p^{y, \sigma| \theta} (y_i^\mathrm{obs}, \sigma_i | \theta). \label{hamil-GMM-25}
\end{align}
Then the Hamiltonian of the GMM for $\Sigma = \{ \sigma_i \}_{i=1}^N$ with $Y = Y_i^\mathrm{obs}$ is given by $H_\mathrm{cl}^{\Sigma| \theta} = \sum_{i=1}^N H_\mathrm{cl}^{\sigma_i| \theta}$.
Using Eq.~\eqref{classical-hamiltonian-10-02}, we can also define the quantum representation of $H_\mathrm{cl}^{\Sigma| \theta}$ as $\hat{H}_\mathrm{cl}^{\Sigma| \theta}$.

To introduce quantum fluctuations into $\hat{H}_\mathrm{cl}^{\Sigma| \theta}$, a non-commutative term $\hat{H}_\mathrm{qu}^{\Sigma} = \sum_{i=1}^N \hat{H}_\mathrm{qu}^{\sigma_i}$ that satisfies Eq.~\eqref{quantization-condition-01} should be added.
In this Letter, we adopt
\begin{align}
\hat{H}_\mathrm{qu}^{\sigma_i} &= \bigg( \overset{i-1}{\underset{j = 1}{\otimes}} \hat{I}^{\sigma_j} \bigg) \otimes \Vast( \sum_{\substack{k = 1, \dots, K, \\ l = k \pm 1}} \Ket{ \sigma_i = l } \Bra{ \sigma_i = k } \Vast) \nonumber \\
& \quad \otimes \bigg( \overset{N}{\underset{j = i+1}{\otimes}} \hat{I}^{\sigma_j} \bigg) \otimes \hat{I}^\theta, \label{non-commutative-01}
\end{align}
where $\Ket{ \sigma_i = 0 } = \Ket{ \sigma_i = K }$ and $\Ket{ \sigma_i = K + 1 } = \Ket{ \sigma_i = 1 }$.
We note that the form of $\hat{H}_\mathrm{qu}^{\sigma_i}$ is not limited to the above definition and has arbitrariness in general.

% \input{paper-dqavb-spin-07-01-result.tex}
% !TEX root = ./paper-dqavb-spin-00-01-main.tex
% \section{Numerical simulation} \label{numerical-01}

\begin{figure}[t]
\begin{subfigure}[t]{0.22\textwidth}
\centering
\includegraphics[scale=0.30]{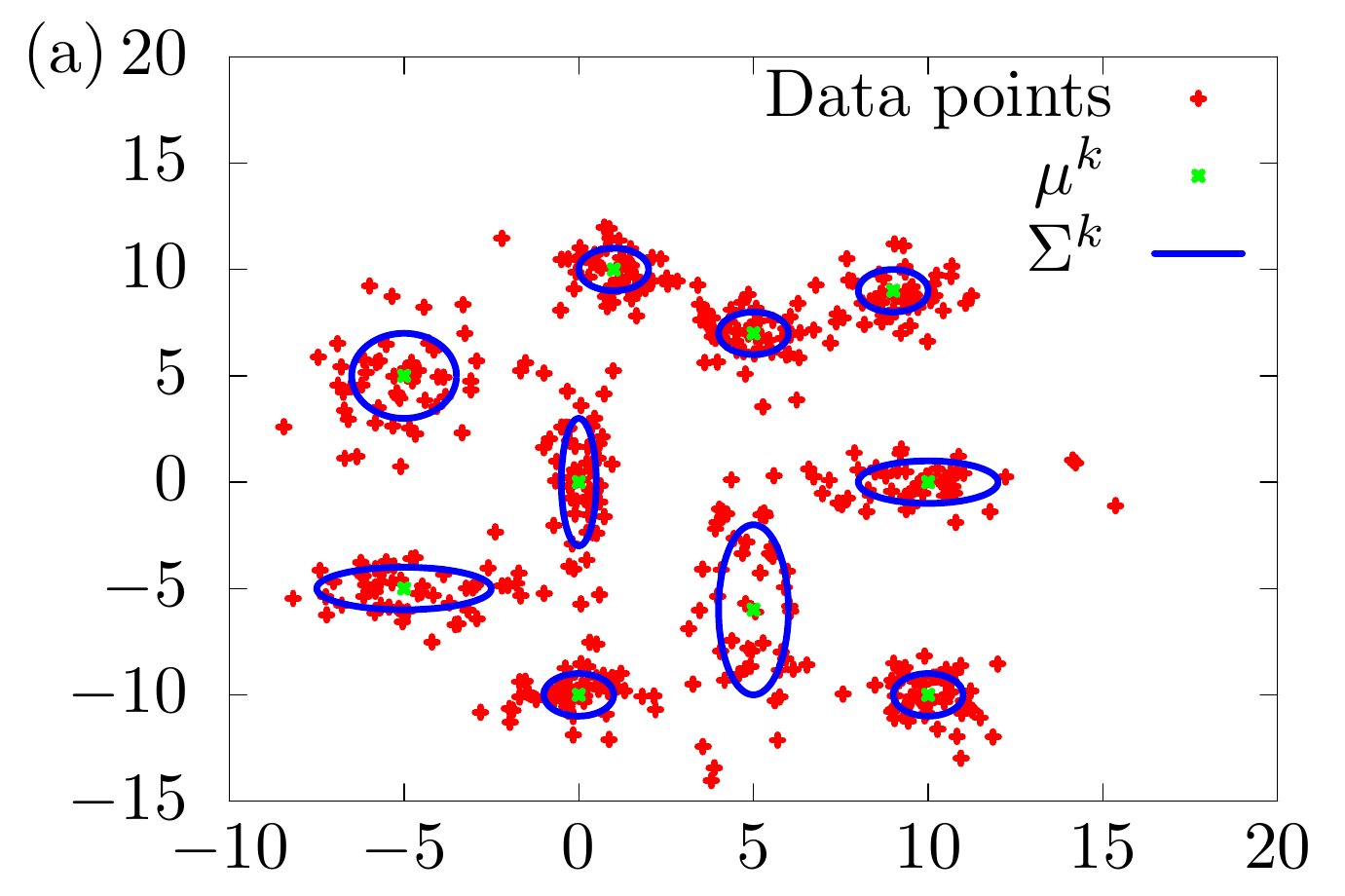}
\end{subfigure}
\begin{subfigure}[t]{0.22\textwidth}
\centering
\includegraphics[scale=0.30]{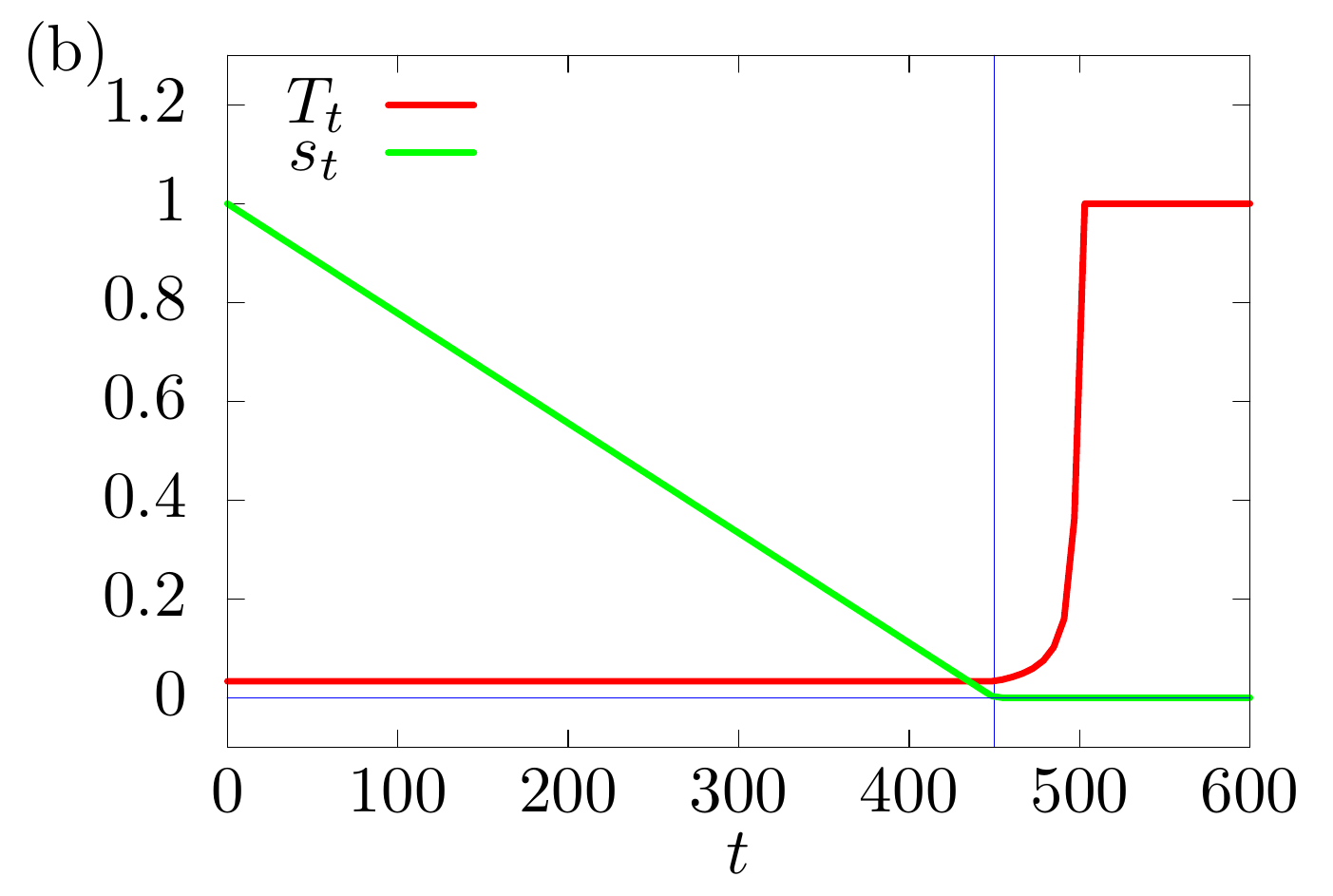}
\end{subfigure}
\caption{(a) Data set generated by 10 Gaussian functions ($K_\mathrm{gen} = 10$). The means and covaricances of Gaussians are depicted by green crosses and blue lines, respectively. (b) Annealing schedules of QAVB. The red line represents $T_t = 1 / \beta_t$ with $\beta_0 = 30.0$, and green lines depict $s_t$ with $s_0 = 1.0$. We set $\tau_\mathrm{QA1} = 450$ and $\tau_\mathrm{QA2} = 500$.}
\label{numerical-01-01}
\end{figure}

\textit{Numerical setup and results.}---
We assess the performances of three algorithms: QAVB, VB, and SAVB.
In this numerical simulation, we use the data set shown in Fig.~\ref{numerical-01-01}(a).
The number of Gaussian mixtures of the generating model $K_\mathrm{gen}$ is 10.
The means and covaricances of Gaussians are depicted by green crosses and blue lines in Fig.~\ref{numerical-01-01}(a), respectively.

There are many candidates for annealing schedules; so, we limit ourselves to some annealing schedules as follows.
Let $\beta_t$ and $s_t$ be $\beta$ and $s$ at the $t$-th iteration, respectively.
For QAVB, we vary $s_t$ and $\beta_t$ as $s_t = s_0 \times \mathrm{max} (1 - t / \tau_\mathrm{QA1}, 0.0)$ and
\begin{align}
\beta_t =
\begin{cases}
\beta_0 & (t \le \tau_\mathrm{QA1}) \\
1 + \frac{ (\beta_0 - 1) (\tau_\mathrm{QA2} - t)}{\tau_\mathrm{QA2} - \tau_\mathrm{QA1}} & (\tau_\mathrm{QA1} \le t \le \tau_\mathrm{QA2}) \\
1.0 & (t \ge \tau_\mathrm{QA2})
\end{cases}, \label{AS-beta-QAVB-01}
\end{align}
respectively, where $s_0$ and $\beta_0$ are initial values of the annealing schedules, $\tau_\mathrm{QA1}$ and $\tau_\mathrm{QA2}$ specify the time scales of the annealing schedules, and $\mathrm{max} (x, y)$ gives the maximum of $x$ and $y$.
To visualize how $T_t \coloneqq 1 / \beta_t$ and $s_t$ behave in the above annealing schedules, we illustrate them in Fig.~\ref{numerical-01-01}(b).
The reason why we adopt the above annealing schedules will be discussed later.
Note that QAVB with $s_0 = 0$ corresponds to SAVB and SAVB with $\beta _0 = 1$ is identical to VB.

We show the numerical results of the three algorithms~\footnote{In the numerical simulation, we used the Dirichlet distribution $\mathcal{D} (\pi | \{ \alpha_\mathrm{pr}^k \}_{k=1}^K)$, Gauss distribution $\prod_{k=1}^K \mathcal{N} (\mu^k | m_\mathrm{pr}^k, (\gamma_\mathrm{pr}^k \Lambda^k)^{-1})$, and the Wishart distribution $\prod_{k=1}^K \mathcal{W} (\Lambda^k | W_\mathrm{pr}^k, \nu_\mathrm{pr}^k)$ for the prior distributions of $\pi$, $\mu$, and $\Lambda$, respectively; that is, $p_\mathrm{pr} (\theta = \{\pi, \mu, \Lambda \})$ $= \mathcal{D} (\pi | \{ \alpha_\mathrm{pr}^k \}_{k=1}^K)$ $\prod_{k=1}^K \mathcal{N} (\mu^k | m_\mathrm{pr}^k, (\gamma_\mathrm{pr}^k \Lambda^k)^{-1})$ $\prod_{k=1}^K \mathcal{W} (\Lambda^k | W_\mathrm{pr}^k, \nu_\mathrm{pr}^k)$. Here, we do not provide the definitions of the three distributions. If the reader is not familiar with them, see Ref.~\cite{Bishop01, Murphy01}. Furthermore, we set $\alpha_\mathrm{pr}^k = 0.001$, $\gamma_\mathrm{pr}^k = 0.001$, $m_\mathrm{pr}^k = \vec{0}$ where $\vec{0}$ is a zero vector, $W_\mathrm{pr}^k = I$ where $I$ is an identity operator, and $\nu_\mathrm{pr}^k = 1$ for each $k$.}.
We set $K = 15$ hereafter.
In Fig.~\ref{numerical-02-01}(a), we first compare QAVB and VB by plotting the estimated number of clusters and the posterior log-likelihood, which is given by
\begin{align}
\mathcal{L} \Big( q^\Sigma (\cdot) q^\theta (\cdot) \Big) &= - \mathrm{KL} \left( q^\Sigma (\cdot) q^\theta (\cdot) \middle\| p^{\Sigma, \theta | Y} (\cdot, \cdot | Y^\mathrm{obs}) \right) \nonumber \\
& \quad + \ln p^Y (Y^\mathrm{obs}). \label{free-energy-xx-01}
\end{align}
To draw Fig.~\ref{numerical-02-01}(a), we ran QAVB and VB 1000 times with randomized initialization.
For QAVB, we set $s_0 = 1.0$, $\beta_0 = 30.0$, $\tau_\mathrm{QA1} = 450$, and $\tau_\mathrm{QA2} = 500$.
Estimates with the same number of clusters and same posterior log-likelihood are plotted at the same point in Fig.~\ref{numerical-02-01}(a).
To count trials with the same estimate, we represent them by error bars along the horizontal axis; thus long lines mean that they are frequently obtained in 1000 trials, while short lines mean that they are rarely obtained.
Furthermore, the lengths of the error bars are normalized to ten for VB and unity for QAVB.
Figure~\ref{numerical-02-01}(a) shows that, while VB can never find it, all the trials of QAVB attain the best posterior log-likelihood.
That is, the success ratio of QAVB is $100.0 \%$.
Next, we show the comparison between QAVB and SAVB in Fig.~\ref{numerical-02-01}(b).
or SAVB, we adopt $\beta_t = 1 + (\beta_0 - 1) \times \mathrm{max} (1 - t / \tau_\mathrm{SA}, 0.0)$, because Eq.~\eqref{AS-beta-QAVB-01} is not an effective one, and we set $\beta_0 = 0.9$ and $\tau_\mathrm{SA} = 500$.
{The length of the error bars for SAVB is also normalized to ten as those for VB.
Figure~\ref{numerical-02-01}(b) also shows that SAVB fails to find the best posterior log likelihood while QAVB finds it~\footnote{Although we have checked different $\beta_0$, SAVB cannot find the best estimate.}.

This numerical simulation shows the surprising superiority of QAVB against VB and SAVB, because only QAVB attains the best posterior log-likelihood.
Furthermore, the computational cost of QAVB scales linearly against the number of data points $N$ and thus QAVB works well even for large $N$.

\begin{figure}[t]
\begin{subfigure}[t]{0.22\textwidth}
\centering
\includegraphics[scale=0.25]{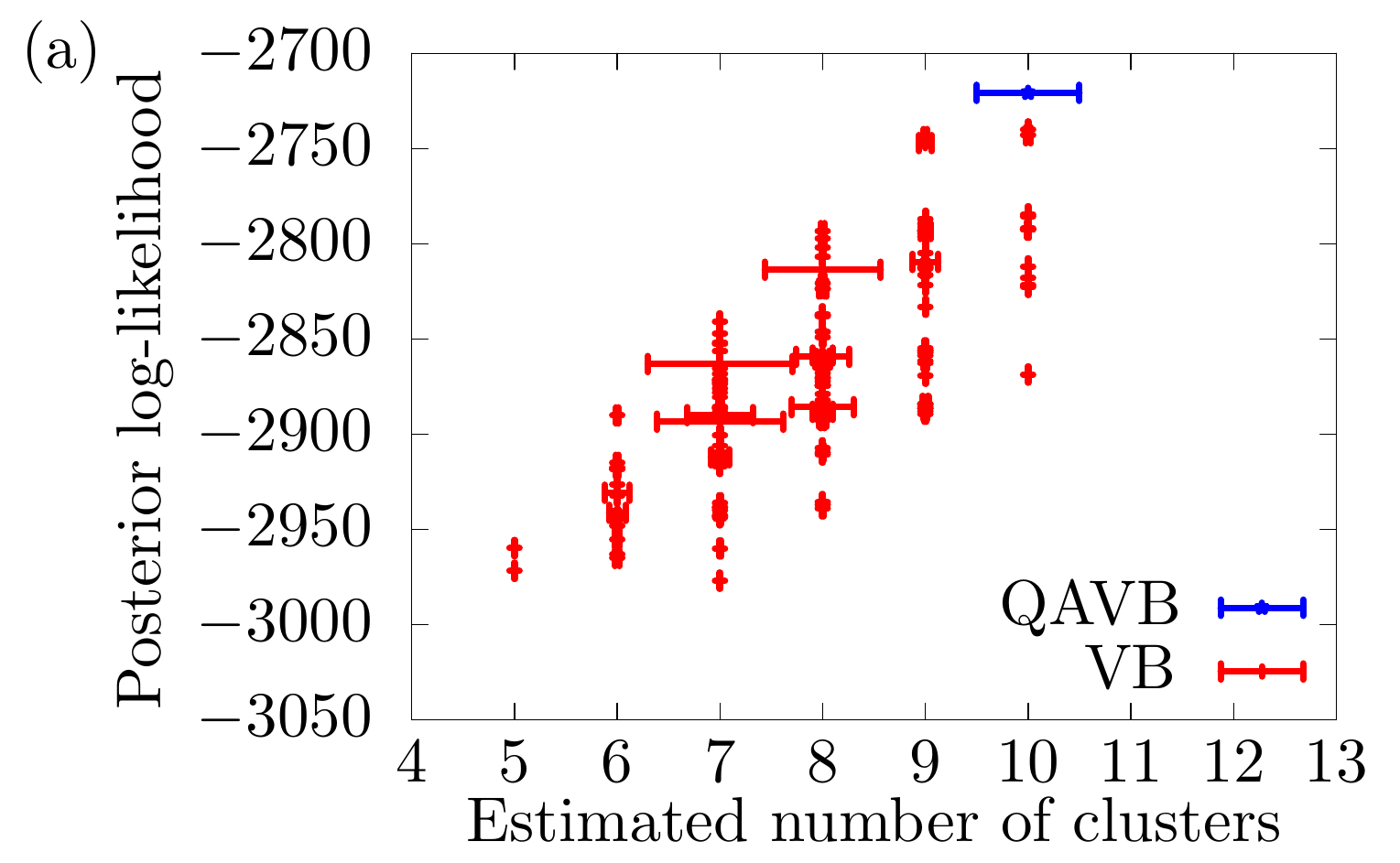}
\end{subfigure}
\begin{subfigure}[t]{0.22\textwidth}
\centering
\includegraphics[scale=0.25]{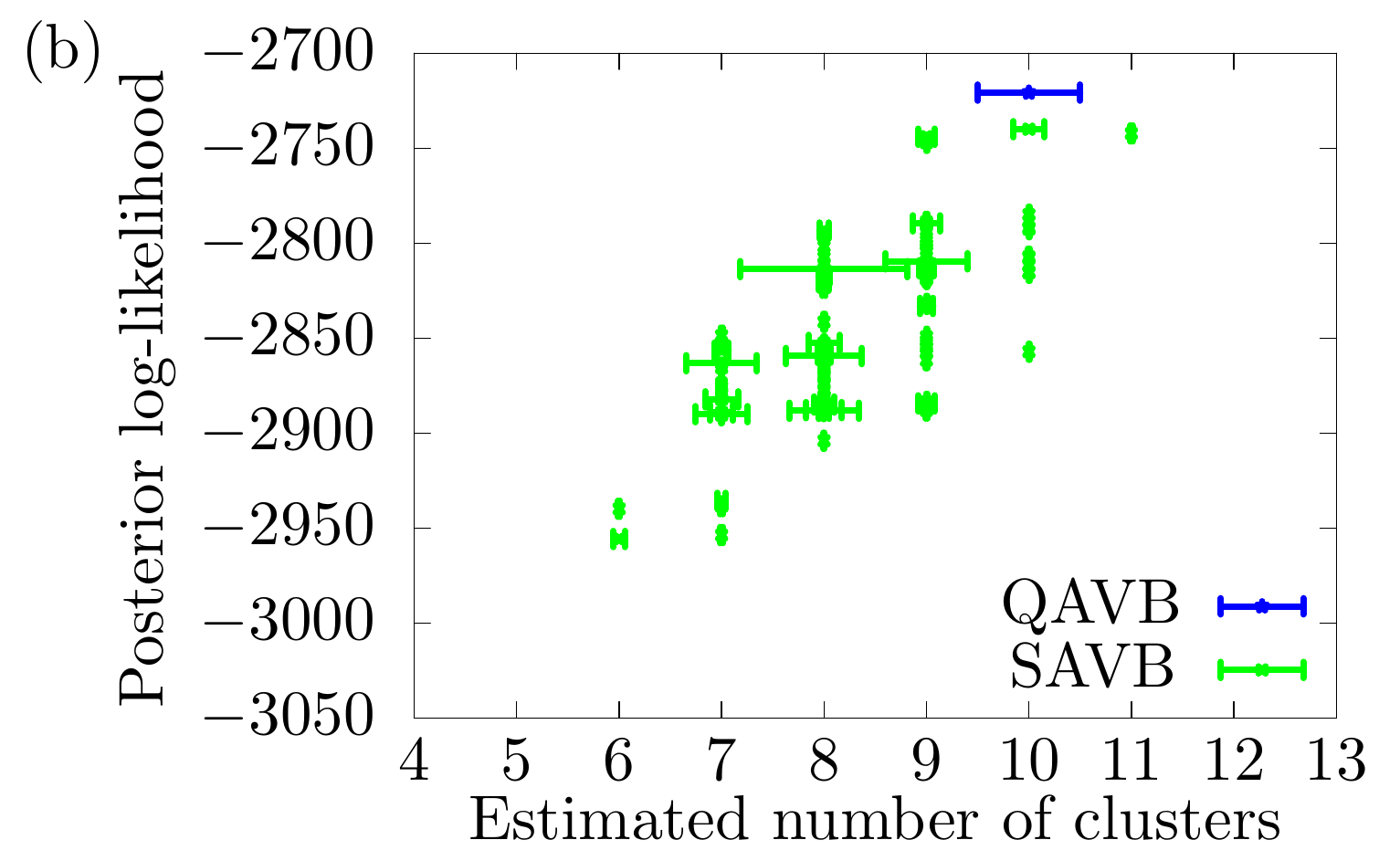}
\end{subfigure}
\caption{(a) Relation between the number of estimated clusters and the posterior log-likelihood of QAVB and VB, and (b) that of QAVB and SAVB. We set $s_0 = 1.0$ and $\beta_0 = 30.0$ for QAVB and $\beta_0 = 0.9$ for SAVB. The horizontal axis represents the number of estimated clusters, while the vertical axis depicts the posterior log-likelihood. The error bars along the horizontal axis represent frequency normalized to ten for VB and SAVB, and to unity for QAVB.}
\label{numerical-02-01}
\end{figure}

%
% \input{paper-dqavb-spin-07-02-disc.tex}
% !TEX root = ./paper-dqavb-spin-00-01-main.tex
% \section{Numerical simulation} \label{numerical-01}

\textit{Discussion.}---
Here, we intuitively discuss the reason why QAVB is superior to VB and SAVB.
First, we consider the first iteration of QAVB in the numerical simulation.
Then, we have
\begin{align}
\hat{\rho}_{1}^{\Sigma}
&= \frac{\exp \left( \mathrm{Tr}_{\theta} \left[ \hat{\rho}_{1}^\theta \Big( - \hat{H}_\mathrm{pr}^\theta - \beta_0 \hat{H}_\mathrm{qu}^\Sigma \Big) \right] \right)}{\mathcal{Z}_\mathrm{MF} (\beta_0, 1.0)}  \\
&\approx \Ket{\mathrm{GS}} \Bra{\mathrm{GS}}, \label{approx-01}
\end{align}
where $\Ket{\mathrm{GS}}$ is the ground state of $\hat{H}_\mathrm{qu}^\Sigma$, and $\mathcal{Z}_\mathrm{MF} (\beta, s)$ is the mean field partition function with $\beta$ and $s$~\footnote{When $s = 1.0$, it holds that $\mathcal{Z}_\mathrm{MF} (\beta, s = 1.0) = \mathcal{Z} (\beta, s = 1.0)$, because Eq.~\eqref{Gibbs-05-01} does not have a term over $\Sigma$ and $\theta$ and then the mean field approximation is exact.}.
Here, we have assumed that $\beta_0$ is sufficiently large and ignored excited states in the approximation~\eqref{approx-01}.
Next, let us turn our attention to the annealing schedules in the numerical simulation, which consists of two parts: $0 \le t \le \tau_\mathrm{QA1}$ and $\tau_\mathrm{QA1} \le t \le \tau_\mathrm{QA2}$.
In the first part, we gradually decrease $s$ to 0 at low temperature.
The estimated state $\hat{\rho}_t^\Sigma \otimes \hat{\rho}_t^\theta$ is considered to keep staying at the mean field ground state of $(1 - s_t) \hat{H}_\mathrm{cl}^{\Sigma| \theta} + s_t \hat{H}_\mathrm{qu}^\Sigma$ during $0 \le t \le \tau_\mathrm{QA1}$, when $s_t$ is changed slowly enough~\footnote{We expect that something like the adiabatic theorem in quantum mechanics would hold during the process of QAVB.}.
If the above consideration holds, at the $\tau_\mathrm{QA1}$-th iteration, $\hat{\rho}_{\tau_\mathrm{QA1}}^\Sigma \otimes \hat{\rho}_{\tau_\mathrm{QA1}}^\theta$ reaches the mean field ground state of $\hat{H}_\mathrm{cl}^{\Sigma| \theta}$.
In the second part, the temperature increases.
In most cases, during the process to increase the temperature of a system, its state relaxes to a unique equilibrium state at $\beta$.
We therefore expect that, during $\tau_\mathrm{QA1} \le t \le \tau_\mathrm{QA2}$, $\hat{\rho}_t^\Sigma \otimes \hat{\rho}_t^\theta$ would transit from the mean field ground state of $\hat{H}_\mathrm{cl}^{\Sigma| \theta}$ to the Gibbs operator that minimizes Eq.~\eqref{free-mean-42} with $\beta = 1.0$ and $s = 0.0$, and we  finally obtain $q_t^\Sigma (\Sigma) q_t^\theta (\theta) = \Braket{\Sigma, \theta | \hat{\rho}_t^\Sigma \otimes \hat{\rho}_t^\theta | \Sigma, \theta}$ that minimize Eq.~\eqref{KL-VB-01}.
In the above discussion, we have used some non-trivial assumptions without proving them mathematically.
Then, a rigorous discussion on the dynamics of QAVB is an issue in the future.

%

% \input{paper-dqavb-spin-08-01-disc.tex}

% \input{paper-dqavb-spin-08-01-conc.tex}
% !TEX root = ./paper-dqavb-spin-00-01-main.tex
% \section{Conclusion} \label{conc-01}

\textit{Conclusion.}---
We have presented QAVB by introducing quantum fluctuations into VB.
After formulating QAVB, we have shown the numerical simulations, which suggest that QAVB is superior to VB and SAVB, and discussed its mechanism.
We consider that our quantization approach for VB can be applied to other algorithms in machine learning and may yield considerable improvements on them.
Thus, we believe that our approach opens the door to a new field spreading over physics and machine learning.

\medskip

\bibliographystyle{apsrev4-1}
\bibliography{paper-dqavb-spin-99-01-bib}

\appendix

% \input{paper-dqavb-spin-11-11-derivation.tex}
% !TEX root = ./paper-dqavb-spin-00-01-supp.tex

\section{A. Functional derivatives of $\mathcal{G}$ with respect to $\hat{\rho}^{\Sigma}$ and $\hat{\rho}^{\theta}$} \label{sm-01-01}

Here, we derive Eq.~\eqref{free-mean-42} in the main text.
By substituting the mean field approximation $\hat{\rho}^{\Sigma, \theta} = \hat{\rho}^\Sigma \otimes \hat{\rho}^{\theta}$ into Eq.~\eqref{KL-DQAVB-01}, we obtain
\begin{align}
&\mathcal{S} \left( \hat{\rho}^\Sigma \otimes \hat{\rho}^{\theta} \middle \| \frac{\hat{f} (\beta, s)}{\mathcal{Z} (\beta, s)} \right) \nonumber \\
& \quad = - \sum_{\Sigma \in S^\Sigma} \sum_{\Sigma' \in S^\Sigma} \int_{\theta \in S^\theta} d\theta \, \int_{\theta' \in S^\theta} d\theta' \, \nonumber \\
& \qquad \quad \times \Big[ \Bra{\Sigma} \otimes \Bra{\theta} \Big] \Big[ \hat{\rho}^\Sigma \otimes \hat{\rho}^{\theta} \Big] \Big[ \Ket{\Sigma'} \otimes \Ket{\theta'} \Big] \nonumber \\
& \qquad \quad \times \Big[ \Bra{\Sigma'} \otimes \Bra{\theta'} \Big] \Big[ \ln \hat{f} (\beta, s) \Big] \Big[ \Ket{\Sigma} \otimes \Ket{\theta} \Big] \nonumber \\
& \qquad + \sum_{\Sigma \in S^\Sigma} \sum_{\Sigma' \in S^\Sigma} \int_{\theta \in S^\theta} d\theta \, \int_{\theta' \in S^\theta} d\theta' \, \nonumber \\
& \qquad \quad \times \Big[ \Bra{\Sigma} \otimes \Bra{\theta} \Big] \Big[ \hat{\rho}^\Sigma \otimes \hat{\rho}^{\theta} \Big] \Big[ \Ket{\Sigma'} \otimes \Ket{\theta'} \Big] \nonumber \\
& \qquad \quad \times \Big[ \Bra{\Sigma'} \otimes \Bra{\theta'} \Big] \Big[ \ln \Big( \hat{\rho}^\Sigma \otimes \hat{\rho}^{\theta} \Big) \Big] \Big[ \Ket{\Sigma} \otimes \Ket{\theta} \Big] \nonumber \\
& \qquad + \ln \mathcal{Z} (\beta, s). \label{free-mean-01}
\end{align}
where $\Ket{\Sigma, \theta} = \Ket{\Sigma} \otimes \Ket{\theta}$.
This expression can be simplified further using the following identities:
\begin{align}
\ln \Big( \hat{\rho}^{a} \otimes \hat{\rho}^b \Big)
& = \Big( \ln \hat{\rho}^{a} \Big) \otimes \hat{I}^b + \hat{I}^{a} \otimes \Big( \ln \hat{\rho}^b \Big), \\
\Big[ \Bra{a} \otimes \Bra{ b } \Big] & \Big[ \hat{\rho}^{a} \otimes \hat{\rho}^{b} \Big] \Big[ \Ket{a'} \otimes \Ket{b'} \Big] \nonumber \\
& = \Braket{a| \hat{\rho}^{a} |a'} \Braket{ b | \hat{\rho}^{b} |b'},
\end{align}
where $\hat{I}^{a}$ and $\hat{I}^b$ are identity operators in the Hilbert spaces spanned by $\Ket{a}$ and $\Ket{b}$, respectively.
Then we obtain
\begin{align}
\mathcal{S} &\left( \hat{\rho}^\Sigma \otimes \hat{\rho}^{\theta} \middle \| \frac{\hat{f} (\beta, s)}{\mathcal{Z} (\beta, s)} \right) \nonumber \\
& = - \sum_{\Sigma \in S^\Sigma} \sum_{\Sigma' \in S^\Sigma} \int_{\theta \in S^\theta} d\theta \, \int_{\theta' \in S^\theta} d\theta' \, \nonumber \\
& \quad \quad \times \Braket{\Sigma | \hat{\rho}^\Sigma | \Sigma'} \Braket{\theta | \hat{\rho}^\theta | \theta'} \nonumber \\
& \quad \quad \times \Big[ \Bra{\Sigma'} \otimes \Bra{\theta'} \Big] \Big[ \ln \hat{f} (\beta, s) \Big] \Big[ \Ket{\Sigma} \otimes \Ket{\theta} \Big] \nonumber \\
& \quad + \sum_{\Sigma \in S^\Sigma} \sum_{\Sigma' \in S^\Sigma} \Bra{\Sigma} \hat{\rho}^\Sigma \Ket{\Sigma'} \Bra{\Sigma'} \ln \hat{\rho}^\Sigma \Ket{\Sigma} \nonumber \\
& \quad + \int_{\theta \in S^\theta} d\theta \, \int_{\theta' \in S^\theta} d\theta' \, \Bra{\theta} \hat{\rho}^{\theta} \Ket{\theta'} \Bra{\theta'} \ln \hat{\rho}^{\theta} \Ket{\theta} \nonumber \\
& \quad + \ln \mathcal{Z} (\beta, s), \label{free-mean-02}
\end{align}
which is identical to Eq.~\eqref{free-mean-42}.

% \input{paper-dqavb-spin-11-12-derivation.tex}
% !TEX root = ./paper-dqavb-spin-00-01-supp.tex
\section{B. Derivation of update equations} \label{sm-01-02}

We derive the update equations of QAVB, Eqs.~\eqref{DQAVB-Estep01} and \eqref{DQAVB-Mstep01}, from Eq.~\eqref{free-mean-42}.
For preparation of the derivation, we prove the following equality:
\begin{align}
\frac{\delta}{\delta \Braket{\Sigma | \hat{\rho} | \Sigma'}} \mathrm{Tr} \Big[ \hat{X} \ln \hat{\rho} \Big] &= \Braket{ \Sigma' | \hat{X} \hat{\rho}^{-1} | \Sigma}, \label{B1-xxx}
\end{align}
for and density operator $\hat{\rho}$ and any $\hat{X}$ that commutes with $\hat{\rho}$.
The proof is as follows.
\begin{proof}
When $\hat{0} \prec \hat{\rho} \prec 2 \hat{I}$,
the definitions of the logarithm and inverse are given by
\begin{align}
\ln \hat{\rho} &\coloneqq \sum_{n = 1}^\infty \frac{(-1)^{n+1}}{n} (\hat{\rho} - \hat{I})^n, \label{B2-xxx} \\
\hat{\rho}^{-1} &\coloneqq \sum_{n = 1}^\infty (-1)^{n+1} (\hat{\rho} - \hat{I})^{n-1}. \label{B3-xxx}
\end{align}
By substituting Eq.~\eqref{B2-xxx} into the left-hand side of Eq.~\eqref{B1-xxx}, we get
\begin{align}
&\frac{\delta}{\delta \Braket{\Sigma | \hat{\rho} | \Sigma'}} \mathrm{Tr} \Big[ \hat{X} \ln \hat{\rho} \Big] \nonumber \\
&= \sum_{n = 1}^\infty \frac{\delta}{\delta \Braket{\Sigma | \hat{\rho} | \Sigma'}} \mathrm{Tr} \bigg[ \hat{X} \frac{(-1)^{n+1}}{n} (\hat{\rho} - \hat{I})^n \bigg]. \label{B4-xxx}
\end{align}

Each term in the summation in Eq.~\eqref{B4-xxx} can be calculated as
\begin{align}
& \frac{\delta}{\delta \Braket{\Sigma | \hat{\rho} | \Sigma'}} \mathrm{Tr} \bigg[ \hat{X} \frac{(-1)^{n+1}}{n} (\hat{\rho} - \hat{I})^n \bigg] \nonumber \\
& \quad = \frac{(-1)^{n+1}}{n} \sum_{i=1}^n \Braket{\Sigma' | (\hat{\rho} - \hat{I})^{n-i} \hat{X} (\hat{\rho} - \hat{I})^{i-1} | \Sigma} \\
& \quad = (-1)^{n+1} \Braket{\Sigma' | \hat{X} (\hat{\rho} - \hat{I})^{n-1} | \Sigma}. \label{B6-xxx}
\end{align}
We have used $[\hat{X}, \hat{\rho}] = 0$ in Eq.~\eqref{B6-xxx}.
By summing Eq.~\eqref{B6-xxx} over $n$, we have
\begin{align}
&\frac{\delta}{\delta \Braket{\Sigma | \hat{\rho} | \Sigma'}} \mathrm{Tr} \Big[ \hat{X} \ln \hat{\rho} \Big] \nonumber \\
& \quad = \sum_{n=1}^\infty (-1)^{n+1} \Braket{\Sigma' | \hat{X} (\hat{\rho} - \hat{I})^{n-1} | \Sigma} \\
& \quad = \Braket{\Sigma' | \hat{X} \hat{\rho}^{-1} | \Sigma}.
\end{align}
Here, we note the definition of $\hat{\rho}^{-1}$, Eq.~\eqref{B3-xxx}.
\end{proof}

Next, by using Eq.~\eqref{B1-xxx}, we derive the update equations of QAVB, Eqs.~\eqref{DQAVB-Estep01} and \eqref{DQAVB-Mstep01}.
The functional derivative of Eq.~\eqref{free-mean-42} with respect to $\Braket{\Sigma | \hat{\rho}^{\Sigma} | \Sigma'}$ under the constraint $\mathrm{Tr}_{\Sigma} \left[ \hat{\rho}^{\Sigma} \right] = 1$ is given by
\begin{align}
&\frac{\delta}{\delta \Braket{\Sigma | \hat{\rho}^{\Sigma} | \Sigma'}} \Bigg[ \mathcal{S} \left( \hat{\rho}^\Sigma \otimes \hat{\rho}^{\theta} \middle \| \frac{\hat{f} (\beta, s)}{\mathcal{Z} (\beta, s)} \right) \nonumber \\
& \qquad \qquad \qquad \qquad \qquad \qquad \qquad - \alpha \left( \mathrm{Tr}_{\Sigma} \left[ \hat{\rho}^{\Sigma} \right] - 1 \right) \Bigg] \nonumber \\
& \quad = - \int_{\theta \in S^\theta} d\theta \, \int_{\theta' \in S^\theta} d\theta' \, \Braket{\theta | \hat{\rho}^\theta | \theta'} \nonumber \\
& \qquad \quad \times \Big[ \Bra{\Sigma'} \otimes \Bra{\theta'} \Big] \Big[ \ln \hat{f} (\beta, s) \Big] \Big[ \Ket{\Sigma} \otimes \Ket{\theta} \Big] \nonumber \\
& \qquad + \Braket{\Sigma'| \ln \hat{\rho}^{\Sigma} | \Sigma} - (\alpha - 1) \Braket{\Sigma' | \hat{I}^{\Sigma} | \Sigma} \\
& \quad = - \Braket{\Sigma' | \mathrm{Tr}_{\theta} \left[ \hat{\rho}^{\theta} \ln \hat{f} (\beta, s) \right] | \Sigma} \nonumber \\
& \qquad + \Braket{\Sigma'| \ln \hat{\rho}^{\Sigma} | \Sigma} - (\alpha - 1) \Braket{\Sigma' | \hat{I}^{\Sigma} | \Sigma}, \label{functional-derivative-01}
\end{align}
where $\alpha$ is a Lagrange multiplier.

By solving
\begin{align}
&\frac{\delta}{\delta \Braket{\Sigma | \hat{\rho}^{\Sigma} | \Sigma'}} \Bigg[ \mathcal{S} \left( \hat{\rho}^\Sigma \otimes \hat{\rho}^{\theta} \middle \| \frac{\hat{f} (\beta, s)}{\mathcal{Z} (\beta, s)} \right) \nonumber \\
& \qquad \qquad \qquad \qquad \qquad - \alpha \left( \mathrm{Tr}_\Sigma \left[ \hat{\rho}^{\Sigma} \right] - 1 \right) \Bigg] = 0,
\end{align}
we obtain
\begin{align}
\Braket{\Sigma'| \ln \hat{\rho}^{\Sigma} | \Sigma} &= \Braket{\Sigma' | \mathrm{Tr}_{\theta} \left[ \hat{\rho}^{\theta} \ln \hat{f} (\beta, s) \right] | \Sigma} \nonumber \\
& \quad + (\alpha - 1) \Braket{\Sigma' | \hat{I}^{\Sigma} | \Sigma}. \label{update-99-01}
\end{align}
Taking into account that $\Ket{\Sigma}$ and $\Bra{\Sigma'}$ are arbitrary vectors, we obtain
\begin{align}
\ln \hat{\rho}^{\Sigma} &= \mathrm{Tr}_{\theta} \left[ \hat{\rho}^{\theta} \ln \hat{f} (\beta, s) \right] + (\alpha - 1) \hat{I}^{\Sigma}. \label{update-99-31}
\end{align}
Hence we have the update equation of $\Sigma$, Eq.~\eqref{DQAVB-Estep01}, where $\alpha$ contributes as a normalization factor.
On the other hand, by using the same procedure,
we obtain the update equations of $\theta$ as Eq.~\eqref{DQAVB-Mstep01}.

\end{document}